# Deep Learning for Target Classification from SAR Imagery

## Data Augmentation and Translation Invariance

Hidetoshi FURUKAWA[†]

† Toshiba Infrastructure Systems & Solutions Corporation
1 Komukaitoshiba-cho, Saiwai-ku, Kawasaki-shi, Kanagawa, 212–8581 Japan
E-mail: †hidetoshi.furukawa@toshiba.co.jp

**Abstract** This report deals with translation invariance of convolutional neural networks (CNNs) for automatic target recognition (ATR) from synthetic aperture radar (SAR) imagery. In particular, the translation invariance of CNNs for SAR ATR represents the robustness against misalignment of target chips extracted from SAR images. To understand the translation invariance of the CNNs, we trained CNNs which classify the target chips from the MSTAR into the ten classes under the condition of with and without data augmentation, and then visualized the translation invariance of the CNNs. According to our results, even if we use a deep residual network, the translation invariance of the CNN without data augmentation using the aligned images such as the MSTAR target chips is not so large. A more important factor of translation invariance is the use of augmented training data. Furthermore, our CNN using augmented training data achieved a state-of-the-art classification accuracy of 99.6%. These results show an importance of domain-specific data augmentation.

## 1. Introduction

Deep learning, especially convolutional neural networks (CNNs) [1]–[5], improve image recognition performance. To reduce overfitting, AlexNet [1] uses data augmentation and dropout. As data augmentation, AlexNet uses the cropped patches of $224 \times 224$ from the images of $256 \times 256$ pixels in the training phase and averages the prediction of 10-crop in the testing phase.

In automatic target recognition (ATR) from synthetic aperture radar (SAR) imagery, CNNs [6]–[10] has been proposed to classify the SAR images from the Moving and Stationary Target Acquisition and Recognition (MSTAR) public release data set [11]. Among them, the CNNs [8]–[10] use data augmentation. The all-convolutional networks [9] use the cropped patches of $88 \times 88$ from the SAR images of $128 \times 128$ pixels in the training phase as data augmentation.

To understand CNNs, as an analysis of the vertical translation, scale and rotation invariance, the Euclid distance between feature vectors from original and transformed images, and the probability of the correct label for each image are visualized [2]. Also, due to the visualization of translation invariance, the translation-sensitivity map which converted the Euclid distance into a map has been proposed [12].

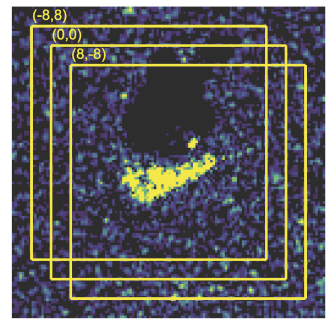

Fig. 1 Illustration of translation $(\Delta x, \Delta y)$. How does the classification accuracy of CNN change with translation?

In SAR ATR, the translation invariance of CNNs represents the robustness against misalignment of target chips extracted from SAR images. Hence, the histogram of accuracy corresponding to $x$ and $y$ displacements visualizes the performance of CNNs with data augmentation [10]. However, there is no analysis of CNNs without data augmentation. Thus the relationship between data augmentation and translation invariance or classification accuracy is not well known.

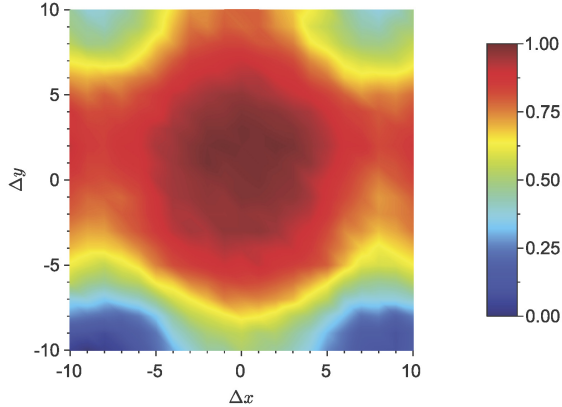

Fig. 2  Accuracy-translation map of the CNN with data augmentation which used the random cropped patches of $96 \times 96$ from the target chips of $104 \times 104$ pixels in the training phase. An accuracy-translation map expresses the classification accuracy of translation $(\Delta x, \Delta y)$.

Table 1  Dataset. The training data contains 3671 target chips (17° depression angle), the test data contains 3203 target chips (15° depression angle) from the MSTAR.

| Class | Training data | Test data |
|---|---|---|
| 2S1 | 299 | 274 |
| BMP2 | 698 | 587 |
| BRDM2 | 298 | 274 |
| BTR60 | 256 | 195 |
| BTR70 | 233 | 196 |
| D7 | 299 | 274 |
| T62 | 299 | 273 |
| T72 | 691 | 582 |
| ZIL131 | 299 | 274 |
| ZSU234 | 299 | 274 |
| Total | 3671 | 3203 |

## 2. Methods

To understand the translation invariance of CNNs, we proposed an accuracy-translation map as a visualization tool, and then visualized the relationship between data augmentation and translation invariance or classification accuracy by using the accuracy-translation map.

### 2.1 Accuracy-translation map

An accuracy-translation map expresses the classification accuracy of translation $(\Delta x, \Delta y)$ where $\Delta x$ and $\Delta y$ denote displacement of $x$ and $y$ direction from the center of the images, respectively. Fig. 2 shows the accuracy-translation map of the CNN with data augmentation.

### 2.2 CNN

CNN for SAR ATR classifies the target images of $96 \times 96$ pixels into the 10 classes. Fig. 3 shows the architecture of the CNN, which based on the 18-layer residual network called ResNet-18 [5]. The CNN contains 17 convolutional layers and one fully-connected (FC) layers. The filter size of the first convolution is $5 \times 5$. The size of other convolutions is $3 \times 3$ like VGG networks [3]. Batch normalization [13] is applied after each convolution and before activation. The activation function of all convolutions uses rectified linear unit (ReLU) [14]. Dropout [15] is not applied.

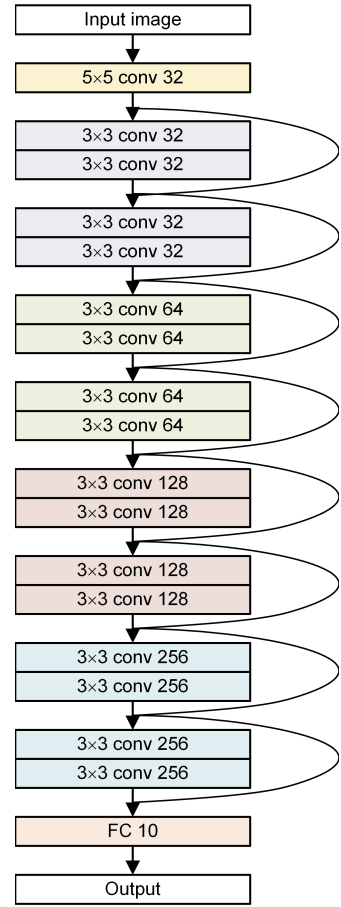

Fig. 3  Architecture of the CNN for SAR ATR. The CNN based on the deep residual network called ResNet-18.

### 2.3 Dataset

For the CNN training and testing, we used the ten classes data shown in Table 1 from the MSTAR [11]. The dataset contains 3671 target chips with a depression angle of 17° for the training and 3203 target chips with a depression angle of 15° for the testing as in [6], [7].

### 2.4 Data augmentation in the training phase

To clarify the relationship between data augmentation and translation invariance or classification accuracy, we trained the CNN with and without data augmentation. The CNN without data augmentation used the center cropped patches of $96 \times 96$ pixels from the target chips in the training phase. The CNN with data augmentation used the random cropped patches of $96 \times 96$ from the target chips of $100 \times 100$ pixels in the training phase.

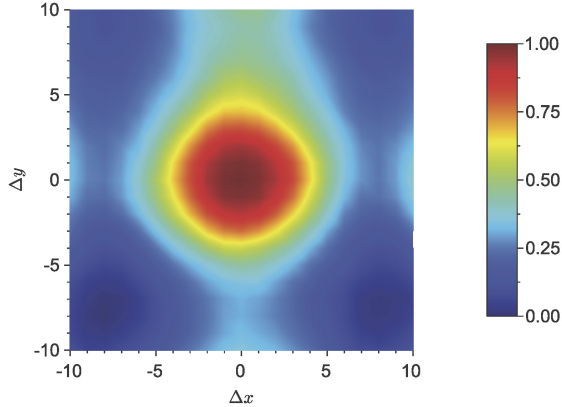 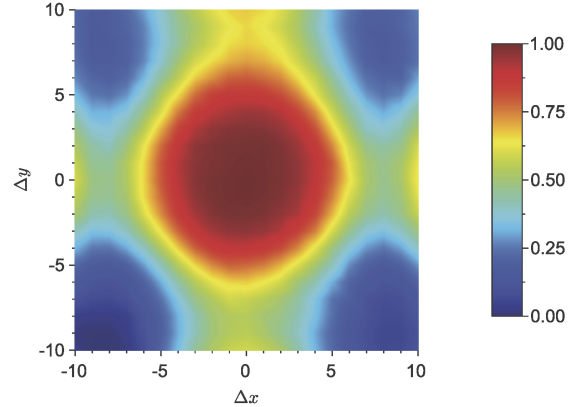

Fig. 4  Accuracy-translation map of the CNN without data augmentation which used the center cropped patches of 96×96 pixels from the target chips in the training phase.

Fig. 5  Accuracy-translation map of the CNN with data augmentation which used the random cropped patches of 96 × 96 from the target chips of 100 × 100 pixels in the training phase.

## 3.　Results

### 3.1　Classification accuracy of the center cropped test data

First, we show the results of classification accuracy of the center cropped test data.

The classification accuracy of the CNN using non-augmented training data is 98.75% (3163/3203). Table 2 shows the confusion matrix of the CNN using non-augmented training data. Each row in the confusion matrix represents the actual target class, and each column denotes the class predicted by the CNN.

The classification accuracy of the CNN using augmented training data is 99.56% (3189/3203). Table 3 shows the confusion matrix of the CNN using augmented training data.

The accuracy of the CNN with data augmentation is higher than the CNN without data augmentation, the CNN using augmented training data achieved a state-of-the-art classification accuracy of 99.6%.

### 3.2　Accuracy-translation map

Then, we show the classification accuracy of translated test data as the accuracy-translation map.

Fig. 4 shows the accuracy-translation map of the CNN using non-augmented training data. The classification accuracy is high in the center $(0, 0)$ with no translation in both $x$ and $y$ direction, and the accuracy decreases as the distance $r = \sqrt{\Delta x^2 + \Delta y^2}$ from the center increases.

Fig. 5 shows the accuracy-translation map of the CNN using augmented training data. The classification accuracy is higher in the range of $-2 \leq \Delta x, \Delta y \leq 2$ where applied random crop in the training phase as data augmentation.

Fig. 6 shows the mean image of the training data. The average image for the target chips indicates that the target located at the center of the target chips. In this case, the effect of the network architecture such as pooling operation and the filter size of convolutions is smaller than augmented training data. Even if we use a deep residual network, the translation invariance of the CNN without data augmentation using the aligned images such as the MSTAR target chips is not so large. It is necessary to displace the target position such as random cropping artificially.

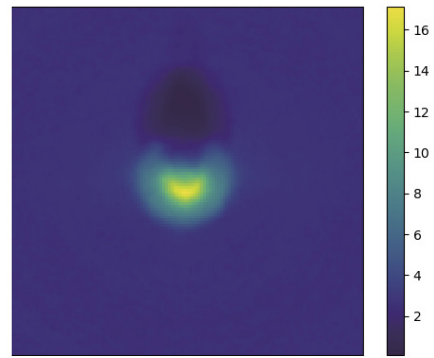

Fig. 6  Mean image of the training data. The mean image of the target chips indicates that the target aligned at the center of the target chips.

### 3.3　Accuracy-translation plot

Finally, we show the classification accuracy of translated test data as an accuracy-translation plot. Fig. 7 shows the accuracy-translation plot of the CNN using non-augmented training data. When $\Delta x$ or $\Delta y$ is ±3, the classification accuracy is less than 90%. Fig. 8 shows the accuracy-translation plot of the CNN using augmented training data. When $\Delta x$ or $\Delta y$ is ±3, the classification accuracy is about 98%.

Table 2  Confusion matrix of the CNN using non-augmented training data. The classification accuracy of the center cropped test data is 98.75%.

| Class | 2S1 | BMP2 | BRDM2 | BTR60 | BTR70 | D7 | T62 | T72 | ZIL131 | ZSU234 | Accuracy(%) |
|---|---|---|---|---|---|---|---|---|---|---|---|
| 2S1 | 268 | 0 | 2 | 0 | 0 | 0 | 3 | 0 | 1 | 0 | 97.81 |
| BMP2 | 0 | 584 | 0 | 0 | 2 | 0 | 0 | 1 | 0 | 0 | 99.49 |
| BRDM2 | 0 | 0 | 268 | 0 | 0 | 0 | 0 | 0 | 6 | 0 | 97.81 |
| BTR60 | 3 | 0 | 4 | 187 | 1 | 0 | 0 | 0 | 0 | 0 | 95.90 |
| BTR70 | 1 | 0 | 0 | 0 | 195 | 0 | 0 | 0 | 0 | 0 | 99.49 |
| D7 | 0 | 0 | 0 | 0 | 0 | 272 | 1 | 0 | 0 | 1 | 99.27 |
| T62 | 0 | 0 | 0 | 1 | 0 | 0 | 264 | 4 | 1 | 3 | 96.70 |
| T72 | 0 | 0 | 0 | 0 | 0 | 0 | 1 | 581 | 0 | 0 | 99.83 |
| ZIL131 | 0 | 0 | 0 | 0 | 0 | 0 | 0 | 0 | 273 | 1 | 99.64 |
| ZSU234 | 0 | 0 | 0 | 0 | 0 | 2 | 0 | 0 | 1 | 271 | 98.91 |
| Total | | | | | | | | | | | 98.75 |

Table 3  Confusion matrix of the CNN using augmented training data. The classification accuracy of the center cropped test data is 99.56%.

| Class | 2S1 | BMP2 | BRDM2 | BTR60 | BTR70 | D7 | T62 | T72 | ZIL131 | ZSU234 | Accuracy(%) |
|---|---|---|---|---|---|---|---|---|---|---|---|
| 2S1 | 273 | 0 | 0 | 0 | 1 | 0 | 0 | 0 | 0 | 0 | 99.64 |
| BMP2 | 0 | 587 | 0 | 0 | 0 | 0 | 0 | 0 | 0 | 0 | 100.00 |
| BRDM2 | 0 | 0 | 273 | 0 | 0 | 0 | 0 | 0 | 1 | 0 | 99.64 |
| BTR60 | 0 | 0 | 5 | 188 | 0 | 1 | 0 | 0 | 0 | 1 | 96.41 |
| BTR70 | 0 | 0 | 1 | 0 | 195 | 0 | 0 | 0 | 0 | 0 | 99.49 |
| D7 | 0 | 0 | 0 | 0 | 0 | 274 | 0 | 0 | 0 | 0 | 100.00 |
| T62 | 0 | 0 | 0 | 0 | 0 | 0 | 269 | 3 | 0 | 1 | 98.53 |
| T72 | 0 | 0 | 0 | 0 | 0 | 0 | 0 | 582 | 0 | 0 | 100.00 |
| ZIL131 | 0 | 0 | 0 | 0 | 0 | 0 | 0 | 0 | 274 | 0 | 100.00 |
| ZSU234 | 0 | 0 | 0 | 0 | 0 | 0 | 0 | 0 | 0 | 274 | 100.00 |
| Total | | | | | | | | | | | 99.56 |

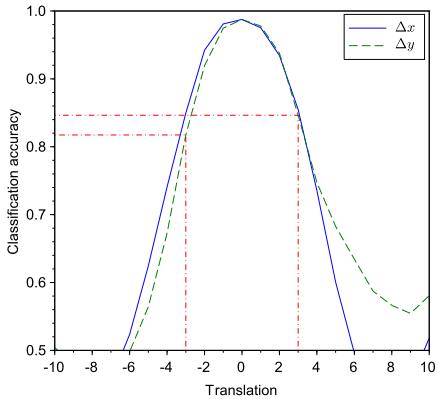

Fig. 7  Accuracy-translation plot of the CNN using non-augmented training data.

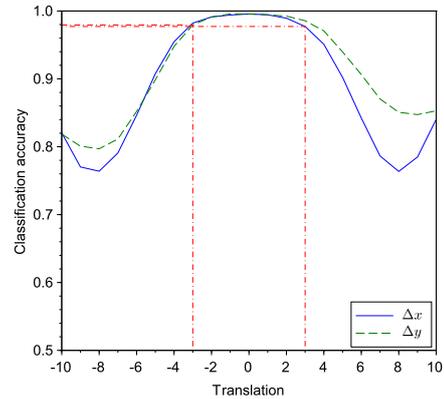

Fig. 8  Accuracy-translation plot of the CNN using augmented training data.

## 4. Conclusion

To understand the translation invariance of CNNs for SAR ATR, we trained deep residual networks which classify the target chips from the MSTAR into the 10 classes under the condition of with and without data augmentation, and then visualized the translation invariance of the CNNs by using the accuracy-translation map we proposed.

According to Fig. 4 and Fig. 5, even if we use a deep residual network, the translation invariance of the CNN without data augmentation using the aligned images such as the MSTAR target chips is not so large. A more important factor of translation invariance is the use of augmented training data.

Furthermore, on the testing using center cropped data, the classification accuracy of the CNN with data augmentation is higher than the CNN without data augmentation, the CNN using augmented training data achieved a state-of-the-art classification accuracy of 99.6%. These results show an importance of the domain-specific data augmentation.